\documentclass{article} 
\usepackage{iclr2027_conference,times}

\usepackage{amsmath,amsfonts,bm}

\def\eqref#1{equation~\ref{#1}}

\def\1{\bm{1}}

\DeclareMathAlphabet{\mathsfit}{\encodingdefault}{\sfdefault}{m}{sl}
\SetMathAlphabet{\mathsfit}{bold}{\encodingdefault}{\sfdefault}{bx}{n}

\usepackage{hyperref}
\usepackage{booktabs}
\usepackage{url}
\usepackage{graphicx}
\usepackage{subcaption}
\usepackage[most]{tcolorbox}
\usepackage[most]{tcolorbox}
\usepackage{listings}
\tcbuselibrary{listings,breakable}
\newtcblisting{promptbox}[1][]{
    enhanced,
    breakable,
    listing only,
    listing engine=listings,
    colback=gray!3,
    colframe=gray!50,
    boxrule=0.5pt,
    arc=2pt,
    left=5pt,
    right=5pt,
    top=5pt,
    bottom=5pt,
    title={#1},
    fonttitle=\bfseries,
    listing options={
        basicstyle=\ttfamily\small,
        breaklines=true,
        breakatwhitespace=true,
        columns=fullflexible,
        keepspaces=true,
        showstringspaces=false,
        literate={—}{{---}}1
    }
}
\newcommand{\method}{\textsc{SumOne}}
\title{\method: A Simple and Efficient Summarizer for Plug-and-Play Agent Memory}

\author{Jizhe Zhang \quad Ang Li \quad Chenheng Zhang \quad Yisen Wang\thanks{Corresponding Author: Yisen Wang (yisen.wang@pku.edu.cn).}\\[4pt]
\normalfont\small State Key Lab of General Artificial Intelligence,\\
\normalfont\small School of Intelligence Science and Technology, Peking University}
\begin{document}

\setcounter{footnote}{1} 
\maketitle

\begin{abstract}
Long-horizon language-model agents accumulate increasingly large interaction histories, making context compaction essential for efficient execution. Existing learned compaction methods often rely on downstream task outcomes or solver behavior, coupling training to specific tasks or agents. We introduce SumOne, a standalone summarizer trained to preserve information in agent histories through masked reconstruction. Given a summary of the expired history, a frozen reconstructor recovers randomly masked portions of the original trajectory; reconstruction quality, combined with a length penalty, provides a task-agnostic reward for optimizing the summarizer with GRPO. This yields self-supervision directly from raw trajectories, without reference summaries or downstream task execution. At inference time, only the lightweight summarizer is retained, enabling plug-and-play context compression for otherwise unchanged agents. We conduct a systematic evaluation of \method{} in long-horizon search and software-engineering tasks. SumOne reduces estimated inference cost by approximately 78\% on DeepSearchQA and 75\% on SWE-bench Verified relative to operating with full context, with absolute decreases of 0.04. The same checkpoint also transfers across downstream models and agent harnesses without additional fine-tuning.
\end{abstract}
\section{Introduction}
\label{sec:introduction}

Large language models are increasingly used to build agents that reason over
multiple steps, call tools, and respond to observations from an external
environment \citep{yao2023react, openai2025gpt5, anthropic2026opus5, team2026kimi}. Such agents need access to earlier
interactions: a search agent may have to combine evidence found at different
steps, while a coding agent may need to recall a user constraint, a previous
code change, or the outcome of a failed test. As the task continues, these
interactions accumulate into a long trajectory. Passing the entire trajectory
to the model increases inference cost and eventually reaches its context
limit \citep{liu2024lost, su2024roformer}. Yet keeping every interaction verbatim is often unnecessary. Tool
outputs contain irrelevant details, and the outcome of a completed step may
be more useful than its full sequence of actions. The challenge is therefore
to reduce the context the agent reads without discarding information it may
need later \citep{hu2025memory}.

Existing approaches to agent context management can be roughly organized into three groups. The first uses prompting or simple rules to shorten the context, for example, by discarding older interactions or asking a model to summarize them. Automatic compaction in Codex and Claude Code shows the practical value of this approach 
\citep{openai2026codexloop,anthropicClaudeCodeCosts}. The second group ties memory management to the agent's task or training process. HiAgent summarizes completed subgoals within the agent workflow; FoldGRPO trains agents to operate with compressed tool responses; \citep{hu-etal-2025-hiagent,sun2025foldgrpo}. These methods can adapt memory to the decisions an agent must make, but their design or training depends on a particular task workflow or downstream solver. The third group provides more general memory frameworks. A-MEM \citep{xu2026mem}, for instance, manages information across context and external storage, with explicit decisions about what to retain and retrieve. Such frameworks offer a broader memory interface, but also require the agent to additionally interact with that interface.

Together, these approaches leave room for a memory mechanism that is agnostic to specific tasks, lightweight to use, while maintaining high information density. We propose to train a dedicated small language model that compresses running context of other agents, only using raw trajectories. However, the central challenge would be the training signal. Given a long agent trajectory and its compressed version, how should one judge its completeness and compactness? In this paper, we use masked reconstruction, a classic training objective in self-supervised learning \citep{lewis2020bart, he2022mae}, as a proxy to assess what a summary preserves. Given a summary of trajectory segments, we first mask parts of the original trajectory, and adopt a frozen model to predict the complete history. As a result, the overlap between the original trajectory and the prediction provides a training signal derived from the data itself, without external references or downstream task execution.

\begin{figure}[htbp]
    \centering
    \begin{subfigure}[b]{0.45\textwidth}
        \centering
        \includegraphics[
            width=0.95\linewidth,
            height=6cm,
            trim=0cm 0.0cm 16.5cm 0.0cm,
            clip]{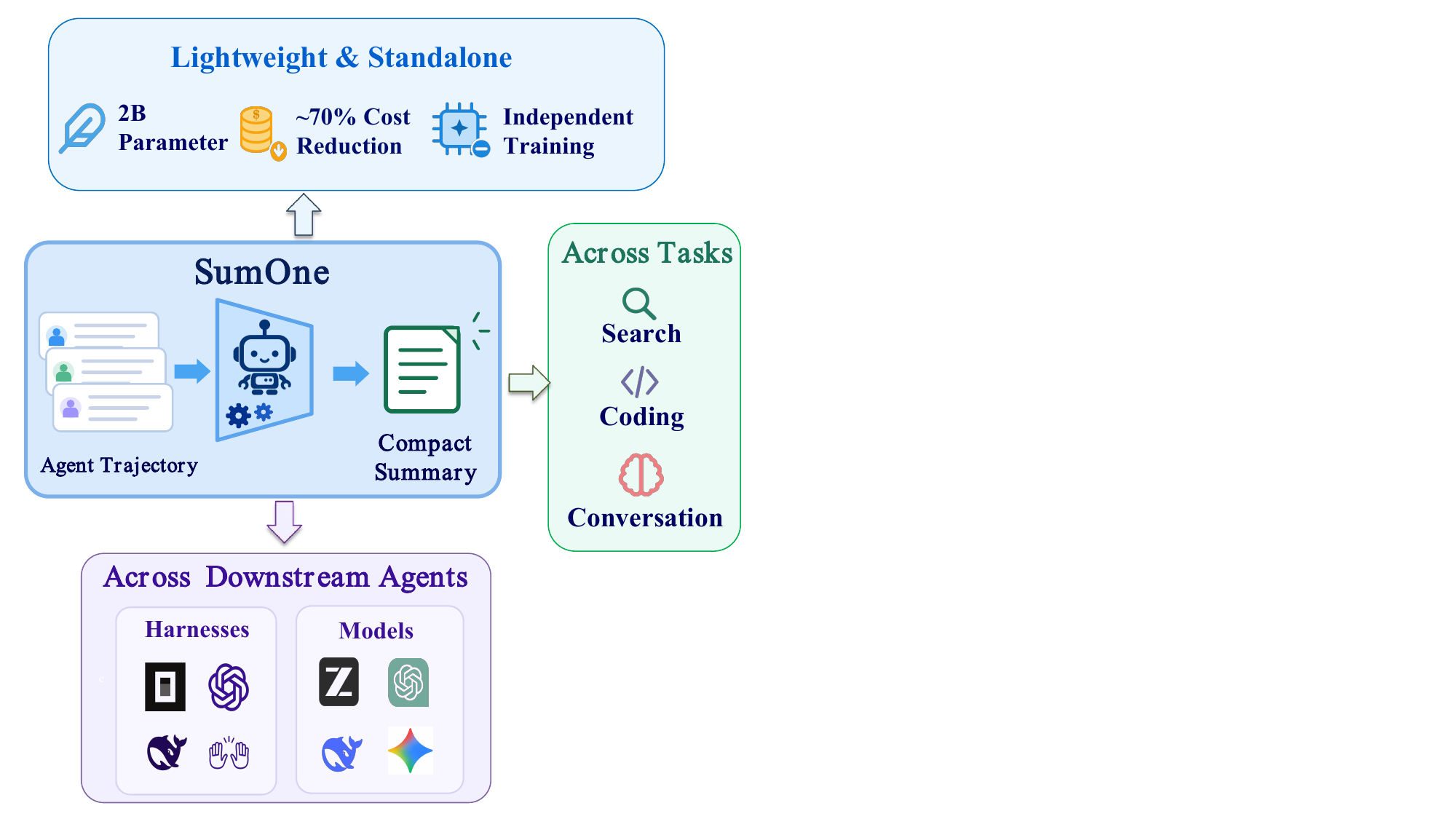}
        \caption{\method{} is a lightweight, standalone memory compressor applicable to diverse agents and tasks.}
        \label{fig:intro_overview}
    \end{subfigure}
    \begin{subfigure}[b]{0.50\textwidth}
        \centering
        \includegraphics[
            width=0.85\linewidth,
            height=6cm,
            trim=0cm 0.0cm 0.5cm 0.0cm,
        ]{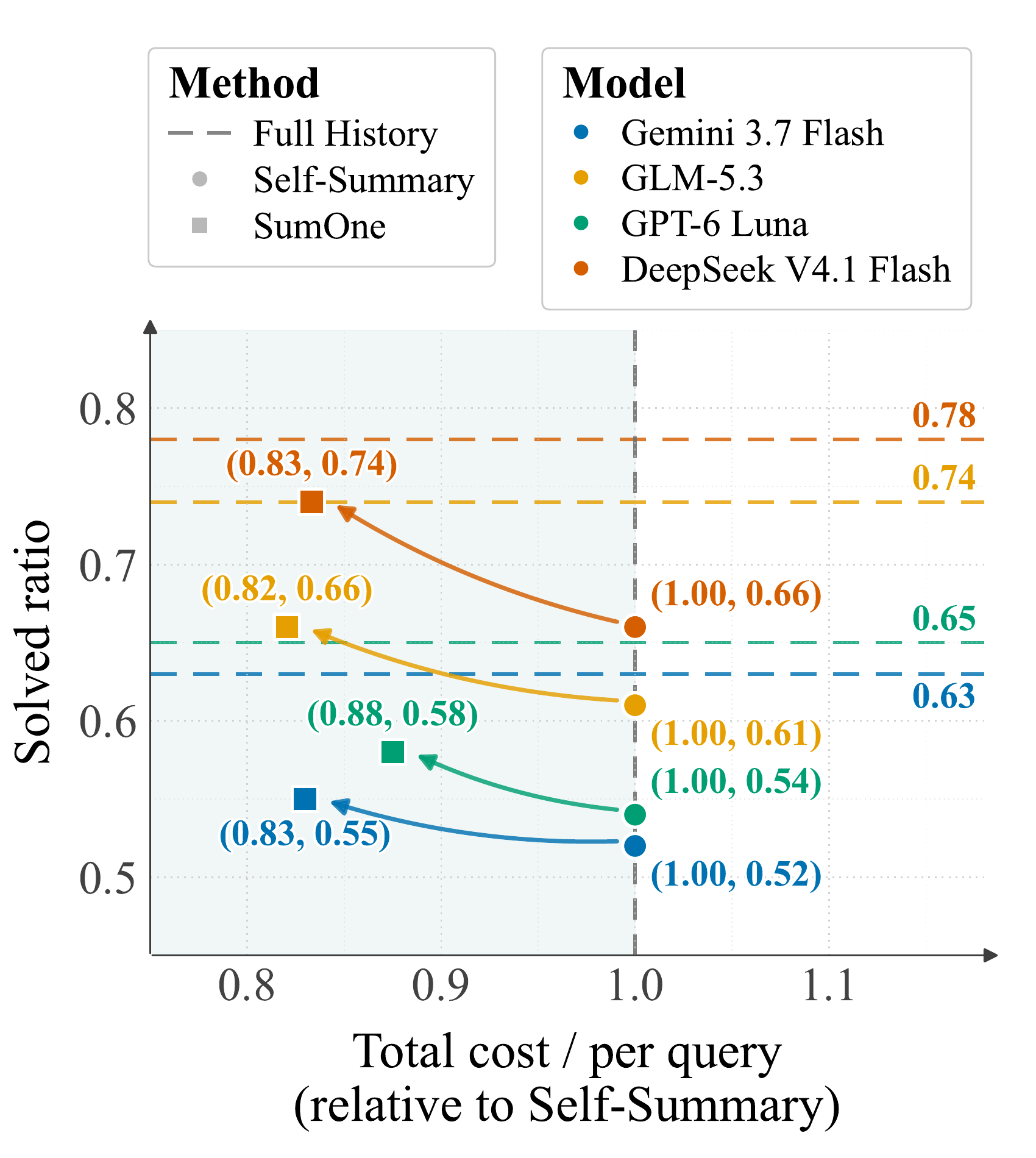}
        \caption{\method{} consistently achieves a better performance--cost trade-off than self-summary}
        \label{fig:intro_tradeoff}
    \end{subfigure}
    \caption{
        \textbf{\method{} is an efficient and scalable memory compressor for long-horizon agents.}
        \textbf{(a)} A single 2B \method{}, trained solely from raw trajectories, can be plugged into different tasks, downstream models,
        and agent harnesses without specific adaptation.
        \textbf{(b)} \method{} consistently improves over self-summary in both task performance and inference cost, yielding a better performance--cost trade-off.  The $y$-axis shows task success rate (resolved rate) on the SWE-bench Verified~\citep{jimenez2024swe} benchmark, while the $x$-axis shows total inference cost per query normalized by that of self-summary.
    }
    \label{fig:intro}
    \vspace{-1mm}
\end{figure}

We implement the above procedure as the reward model in a GRPO-style RL training loop \cite{shao2024deepseekmath}, and collect a training dataset of 5k trajectories primarily sampled from web search and coding agents. The resulting model, which we call \method{}, is a specialized compression model for agentic context with merely 2 billion parameters. As illustrated in Figure~\ref{fig:intro_overview}, \method{} undergoes task-independent training on raw trajectories. It specializes in compressing agent trajectories into textual memory on-the-fly, and proves to be effective across a wide range of settings based on our experiments. 

We evaluate \method{} on long-horizon search and software-engineering tasks, together with agentic memory benchmarks and transfer across downstream agents and harnesses. On DeepSearchQA~\citep{gupta2026deepsearchqa}, \method{} reduces more than 70\% cumulative agent input and 77\% end-to-end task cost per query compared to evaluating with full context, while causing merely 0.04 drop in F1 score. On SWE-bench Verified~\citep{jimenez2024swe}, it achieves a 59\% resolved rate versus 63\%, while reducing input from 4.1M to 1.09M tokens per task. Compared with agent self-summarization, \method{} performs better on both benchmarks while using notably less end-to-end cost,with Figure~\ref{fig:intro_tradeoff} highlighting the results for SWE-bench Verified. It also achieves the strongest results among compression methods on LongMemEval-S~\citep{wu2024longmemeval} and transfers across multiple downstream LLMs and agent harnesses without additional fine-tuning. 

Our broader aim is to make memory management a capability of a dedicated module that can serve different agents, similar to the role that hippocampus plays in human brains \citep{mcclelland1995complementary}. \method{} addresses one part of this problem: learning to write compact memory from raw interaction histories, with supervision obtained from the data itself. To summarize, the advantages of \method{} are as follows:

\begin{itemize}
\item \textbf{Self-supervised training.}
\method{} learns directly from raw agent trajectories, using masked reconstruction to derive its training signal without human annotations, external reference summaries, or downstream task execution.

\item \textbf{Efficient inference.}
\method{} is a lightweight 2B-parameter model specialized for context compression. By replacing inference over an ever-growing, append-only context with a shorter, bounded context maintained by \method{}, it reduces end-to-end task cost by 70\% on the DeepSearch QA and SWE-Bench-Verfied benchmarks, while incurring a modest performance drop of 0.04.

\item \textbf{Plug-and-play deployment.} 
\method{} is trained independently of any particular task, agent model, or harness. Our empirical study demonstrates its transferability across 3 domains, 5 downstream LLMs including GPT-6 \citep{openai2026gpt6luna}, and 4 coding agent harnesses without extra fine-tuning.

\end{itemize}

\section{Related Work}
\label{sec:related-work}

We position our paper in the field of agentic memory, and it is especially related to studies trying to compact running context of agents for efficiency. Our core training algorithm involves mask reconstruction, a classic self-supervised learning objective. Therefore, we briefly review the two research areas below.

\paragraph{Agent memory and context compaction.}
Tool-using agents accumulate long histories of actions and observations~\citep{yao2023react}, motivating mechanisms for retaining useful information under limited context budgets. Existing approaches can be broadly grouped into three categories. \textbf{First}, input-level compression reduces context by selecting, removing, or masking parts of the original input, rather than synthesizing a new memory representation, as in prompt compression~\citep{jiang2023llmlingua,pan2024llmlingua2} and
observation masking~\citep{lindenbauer2025complexity}. \textbf{Second}, agent-managed compaction incorporates memory maintenance into the agent workflow, such as subgoal summarization in HiAgent~\citep{hu-etal-2025-hiagent} and learned context folding~\citep{sun2025foldgrpo,ye2025agentfold}. \textbf{Third}, standalone summarization methods compress interaction histories into textual memory for subsequent agent execution. Methods~\citep{wu2025resum,kang2025acon,zhang2026comem} in this category 
typically tie the learning or evaluation of summaries to downstream task outcomes or solver behavior. \method{} also adopts a standalone summarizer, but learns it with a task-agnostic reconstruction objective.

\paragraph{Reconstruction as a learning signal.}
Reconstruction has long provided self-supervision for learning information-preserving representations. Denoising autoencoders~\citep{vincent2008dae},
 masked language modeling~\citep{devlin2019bert}, sequence denoising~\citep{lewis2020bart}, and masked autoencoders~\citep{he2022mae} learn representations by 
 recovering corrupted or missing input. Closer to our setting, MemTrain~\citep{li2026memtrain} uses masked-entity reconstruction and intermediate recall to train an agent 
 to manage its own memory. In contrast, our method trains a separate textual summarizer, while keeping the downstream agent entirely outside the training loop.

\section{Reinforcing Context Compression with Masked Reconstruction}

\label{sec:method}
We introduce \method{}, a framework for learning compact agent memories without relying on downstream task execution as supervision. Our key idea is to treat summarization as \emph{information-preserving compression}: a good summary should retain information from the expired interaction history that remains useful after the original history is discarded. To train for this property directly, \method{} generates a summary of the expired history (Section~\ref{sec:formulate}), masks part of the original history, and asks a frozen reconstructor to recover the missing content from the summary and retained context (Section~\ref{sec:reconstruction-signal}). Reconstruction quality is then converted into a reward for optimizing the summarizer with GRPO~\citep{shao2024deepseekmath} (Section~\ref{sec:optimization}). At inference time, the reconstruction pipeline is removed, leaving only the lightweight summarizer as a plug-and-play context compression module (Section~\ref{sec:deployment}).

\begin{figure*}[t]
  \centering
  \includegraphics[width=\textwidth]{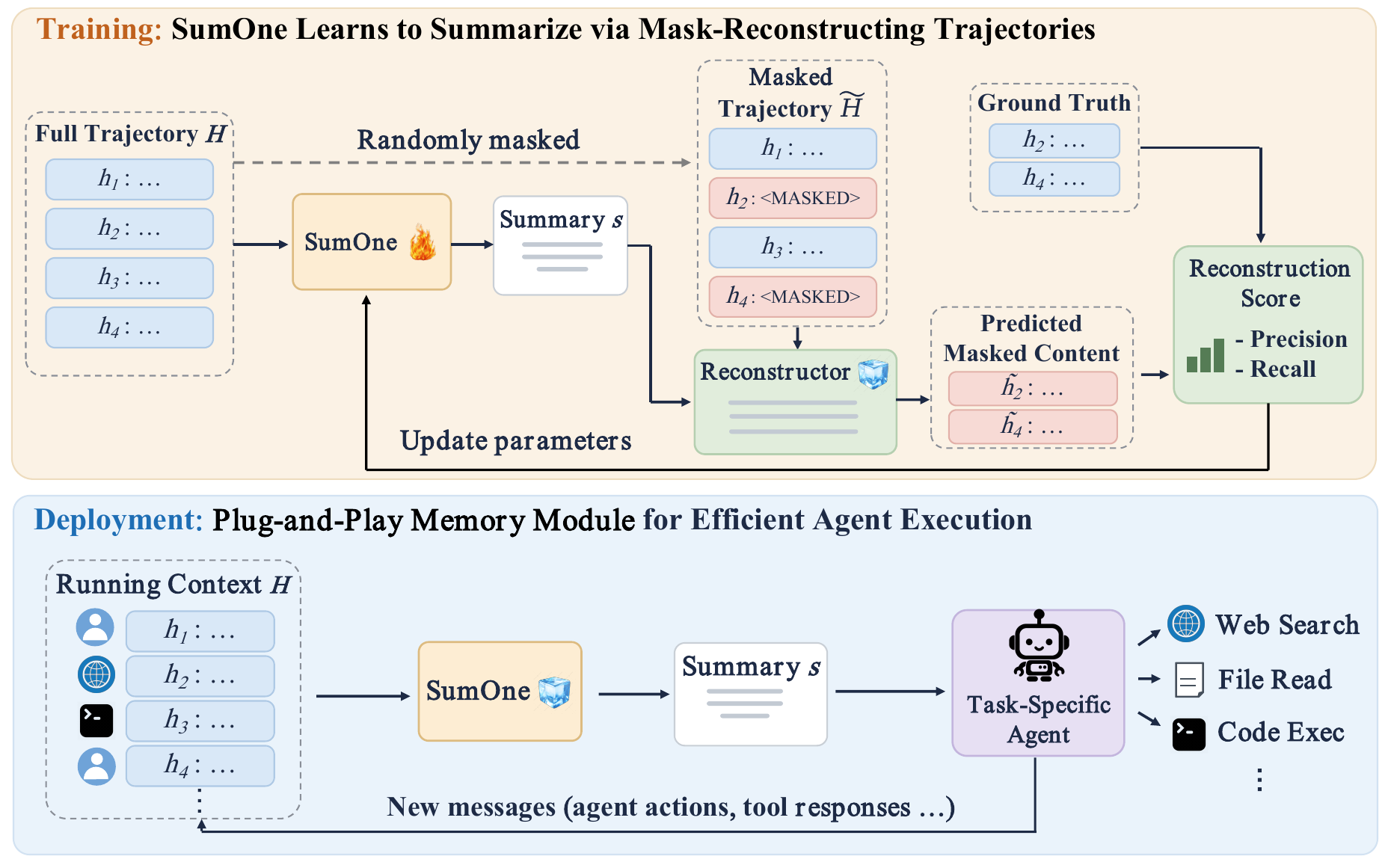}
  \caption{Overview of \method. The summarizer compresses the expired history into a short summary. A frozen reconstructor then attempts to recover masked parts of the original history conditioned on the summary and retained context, and the reconstruction score (with a length penalty) provides the reward to optimize the summarizer.}
  \label{fig:method}
\end{figure*}

\subsection{Formulating the Summary Agent}
\label{sec:formulate}
Let $H$ denote the expired interaction history to be compressed, and let $C$ denote the context retained unchanged during the current compaction step, including the system instructions, task specification, existing memory, and recent interactions. Given $(C,H)$, the summarizer $\pi_\theta$ generates a textual summary
\begin{equation}
S \sim \pi_\theta(\cdot \mid C,H).
\label{eq:summary}
\end{equation}
The objective is to produce a concise $S$ that preserves information in $H$. Rather than supervising $S$ with a reference summary or downstream task reward, \method{} derives its training signal directly from the original history through masked reconstruction.

\subsection{Summary as a Mutual Information Maximizer}
\label{sec:reconstruction-signal}
To measure how much information from $H$ is preserved in $S$, we mask part of the original history and test whether it can be recovered from the summary.
Specifically, we partition $H$ into approximately equal-sized blocks, preferring natural interaction boundaries such as role transitions and tool calls whenever possible. We sample a subset of blocks $\Omega$ and replace them with missing-span markers, producing a partially observed history
\begin{equation}
\widetilde{H}=\operatorname{Mask}(H;\Omega).
\end{equation}
A reconstructor $D$ then predicts the masked content conditioned on the summary and the remaining context:
\begin{equation}
\widehat H_\Omega
\sim
p_D(\cdot\mid C,\widetilde H,S).
\end{equation}

Intuitively, if $S$ preserves information that becomes unavailable after masking $H_\Omega$, conditioning on $S$ should make the missing content easier to recover. Reconstruction quality therefore provides a functional measure of information retained by the summary.

This view also admits an information-theoretic interpretation. Let $H_{\bar\Omega}$ denote the visible portion of the history. The additional information provided by $S$ about the missing content can be characterized by
\begin{equation}
I(S;H_\Omega\mid H_{\bar\Omega},C)
=
\mathcal H(H_\Omega\mid H_{\bar\Omega},C)
-
\mathcal H(H_\Omega\mid S,H_{\bar\Omega},C).
\end{equation}

Since the first term does not depend on the summarizer, a summary that better reduces uncertainty about $H_\Omega$ preserves more information beyond the retained context. \method\ does not estimate this quantity explicitly; reconstruction quality serves as a semantic surrogate.

\subsection{Implementing Masked Reconstruction as a Reward Model in RL}
\label{sec:optimization}

We optimize the summarizer with GRPO based on reconstruction quality. To this end, we first formulate the reconstruction quality as a scalar reward.

An LLM judge evaluates each reconstructed block against its ground-truth content along two dimensions: \emph{precision} and \emph{recall}. \emph{Precision} measures the fraction of the predicted content that is supported by the ground truth, whereas \emph{recall} measures the fraction of the ground-truth content that is successfully recovered. Let $r_i$ and $p_i$ denote the two scores for masked block $h_i$. The overall reconstruction quality is
\begin{equation}
Q(S)=\frac{1}{|\Omega|}\sum_{i\in\Omega}(r_i+p_i).
\end{equation}

Optimizing reconstruction quality alone may favor overly long summaries that preserve most of the original history. We therefore introduce a length penalty and define the final reward as
\begin{equation}
R(S)=Q(S)-\lambda\frac{|S|}{L_{\max}},
\label{eq:reward}
\end{equation}
where $|S|$ denotes the summary length, $L_{\max}$ is the maximum summary budget, and $\lambda$ controls the trade-off between information preservation and compression. This objective captures a compression--fidelity trade-off: $Q(S)$ measures how well information is preserved, while summary length controls the cost of compression.

For each context, we sample $G$ candidate summaries from $\pi_\theta$ and evaluate them using the reconstruction-based reward described above. Following GRPO, we normalize the rewards within each group to obtain a relative advantage for each candidate:
\begin{equation}
A_k=\frac{R(S_k)-\mu_R}{\sigma_R+\epsilon},
\end{equation}
where $\mu_R$ and $\sigma_R$ denote the mean and standard deviation of the group rewards. The summarizer is then optimized with the standard GRPO objective, while the reconstructor remains frozen.

\subsection{Deploying \method{} as a Plug-and-Play Module}
\label{sec:deployment}

Once training is complete, the learned summarizer can be directly paired with a downstream agent for inference. \method\ replaces the expired history $H$ with its summary $S$ from Eq.~\ref{eq:summary}, and the agent continues execution using $(C,S)$. The reconstruction and judge pipeline are no longer required at inference time.

\section{Evaluating \method{}}

\subsection{Experimental Setup}

\paragraph{Training.}
We construct approximately 5K compaction instances at 16K and 32K context lengths, roughly balanced between web-search trajectories from OpenSeeker-v1-Data~\citep{du2026openseeker} and software-engineering trajectories from SWE-smith~\citep{yang2026swe}. For reconstruction, we partition the expired history into roughly 512-token blocks, preserving natural interaction boundaries where possible, and randomly mask 50\% of them. We initialize the SumOne summarizer from Qwen3.5-2B, use a frozen Qwen3.5-9B as the reconstructor, and employ DeepSeek V4.1 Flash to judge content recovery and fabrication. We optimize the summarizer with GRPO for 600 steps using a batch size of 6, a group size of 8, and a learning rate of $1\times10^{-6}$. Additional training details are provided in Appendix~\ref{app:training_details}.

\paragraph{Evaluation.}
Our primary agent benchmarks are DeepSearchQA~\citep{gupta2026deepsearchqa} for web search and SWE-bench Verified~\citep{jimenez2024swebench} for software engineering; we additionally use LongMemEval-S for isolated memory evaluation. Unless otherwise specified, we use Qwen3.5-35B-A3B with OpenCode~\citep{opencode2026} as the downstream agent and harness. Native harness compaction is disabled, and all evaluated methods use a 32K compaction threshold; Full History instead retains the complete trajectory up to the model's native context limit. To test plug-and-play generalization, we additionally evaluate across DeepSeek V4.1 Flash \citep{deepseek2026v41flash}, GPT-6 Luna \citep{openai2026gpt6luna}, GLM-5.3 \citep{zai2026glm53}, and Gemini 3.7 Flash \citep{google2026gemini37flash}, as well as DeepSeek Harness \citep{deepseek2026harness}, OpenHands \citep{wang2025openhands}, and Codex \citep{openai2026codex}.

\paragraph{Baselines and metrics.}
We compare \method{} against five context-management baselines. \textbf{Full History} retains the complete interaction history without compression. \textbf{Recency Truncation}~\citep{ubicloud2026sweqwen3, openai2026agentssdk} discards the oldest interactions once the compaction threshold is reached. \textbf{Self-Summary} uses the downstream agent itself to summarize the expired trajectories. \textbf{Prompt-2B} uses Qwen3.5-2B and summarization prompt as prompt as \method{}, but without any additional training, isolating the effect of our training objective. \textbf{CoMem}~\citep{zhang2026comem} is a learned standalone memory model supervised by downstream agent behavior. All compression methods use the same compaction threshold and output budget whenever applicable. We report downstream task performance, cumulative downstream-agent input tokens, inference cost, average compactions per query, and compression ratio:

\begin{equation}
\mathrm{CR}
=1-\frac{\sum_e |S_e|}{\sum_e |H_e|},    
\end{equation}

where $H_e$ and $S_e$ denote the expired history and its compressed summary at compaction event $e$. Model-specific API prices are provided in Appendix~\ref{app:pricing}.
\begin{table*}[t]
\centering
\small
\setlength{\tabcolsep}{6pt}
\caption{
\textbf{Main results on DeepSearchQA and SWE-bench Verified.}
Score denotes task performance, measured by F1 on DeepSearchQA and the resolved rate on SWE-bench Verified.
C. Ratio, Input Tok., Sum. Cost, Total Cost, and Comp./Q denote the compression ratio,
the number of input tokens consumed by the agent, the summarization cost per query,
the total inference cost per query, and the average number of compactions per query, respectively.
Costs are reported in USD.
}
\label{tab:main_results}

\begin{tabular}{lcccccc}
\toprule

\multicolumn{7}{c}{\textbf{DeepSearchQA}} \\
\midrule
\textbf{Method}
& \textbf{Score $\uparrow$}
& \textbf{C. Ratio $\uparrow$}
& \textbf{Input Tok. $\downarrow$}
& \textbf{Sum. Cost $\downarrow$}
& \textbf{Total Cost $\downarrow$}
& \textbf{Comp./Q $\downarrow$} \\
\midrule

Full History
& 0.52
& --
& 2.83M
& --
& \$0.170
& -- \\

\midrule

Recent Truncation
& 0.16
& --
& 752K
& --
& \$0.047
& 1.24 \\

Self-Summary
& 0.44
& 91\%
& 652K
& \$0.002
& \$0.044
& 1.68 \\

Prompt-2B
& 0.29
& 92\%
& 468K
& \$0.001
& \$0.031
& 1.24 \\

CoMem
& 0.32
& 88\%
& 577K
& \$0.001
& \$0.036
& 1.48 \\

\textbf{\method{}} (ours)
& 0.48
& 90\%
& 591K
& \$0.001
& \$0.038
& 1.57 \\

\bottomrule
\multicolumn{7}{c}{\textbf{SWE-bench Verified}} \\
\midrule

Full History
& 0.63
& --
& 4.1M
& --
& \$0.286
& -- \\

\midrule

Recent Truncation
& 0.46
& --
& 1.27M
& --
& \$0.097
& 2.1 \\

Self-Summary
& 0.56
& 74\%
& 1.05M
& \$0.013
& \$0.082
& 2.7 \\

Prompt-2B
& 0.51
& 77\%
& 0.98M
& \$0.001
& \$0.073
& 2.1 \\

CoMem
& 0.57
& 72\%
& 0.97M
& \$0.001
& \$0.065
& 2.5 \\

\textbf{\method{}} (ours)
& 0.59
& 69\%
& 1.09M
& \$0.001
& \$0.071
& 2.5 \\

\bottomrule
\end{tabular}
\end{table*}

\begin{table*}[t]
\centering

\begin{minipage}[t]{0.52\textwidth}
\centering
\small
\setlength{\tabcolsep}{4pt}

\captionof{table}{
Performance with task-specific RL agents.
}
\label{tab:rl_agent_results}

\begin{tabular}{lcccc}
\toprule
\textbf{Method}
& \textbf{DSQA $\uparrow$}
& \textbf{Cost $\downarrow$}
& \textbf{SWE $\uparrow$}
& \textbf{Cost $\downarrow$} \\
\midrule

Full His. w/o RL
& 0.52
& 0.17
& 0.63
& 0.286 \\

\midrule

Self-Sum.
& 0.57
& 0.052
& 0.65
& 0.091 \\

Prompt-2B
& 0.41
& 0.028
& 0.58
& 0.083 \\

CoMem
& 0.48
& 0.035
& 0.60
& 0.072 \\

\textbf{\method{}}
& 0.55
& 0.039
& 0.68
& 0.077 \\

\bottomrule
\end{tabular}
\end{minipage}
\hfill
\begin{minipage}[t]{0.45\textwidth}
\centering
\small
\setlength{\tabcolsep}{4.5pt}

\captionof{table}{
Cross-harness evaluation with Qwen3.5-35B.
}
\label{tab:cross_harness}

\begin{tabular}{lccc}
\toprule
\textbf{Harness}
& \textbf{Full His.}
& \textbf{Self-Sum.}
& \textbf{\method{}} \\
\midrule

OpenCode
& 0.52
& 0.44
& 0.48 \\

OpenHands
& 0.49
& 0.39
& 0.46 \\

Codex
& 0.55
& 0.47
& 0.47 \\

DS Harness
& 0.50
& 0.41
& 0.42 \\
\bottomrule
\end{tabular}
\end{minipage}

\end{table*}

\subsection{\method{} Substantially Reduces Task Cost with Mild Performance Drop}
\label{sec:main_results}

\textbf{\method{} retains most of the Full History performance at a fraction of the cost.}
Table~\ref{tab:main_results} reports the main results on DeepSearchQA and SWE-bench Verified, using Qwen3.5-35B-A3B as the downstream agent with OpenCode as the agent harness. On DeepSearchQA, \method{} achieves 0.48 F1 compared with 0.52 for Full History, while reducing cumulative agent input from 2.83M to 591K tokens and requiring only 22\% of the total inference cost of Full History. A similar pattern holds on SWE-bench Verified, where \method{} reaches a 0.59 resolved rate compared with 0.63 for Full History, while reducing cumulative input from 4.10M to 1.09M tokens and using only 58\% of the Full History cost. These results show that \method{} preserves most of the utility of the full interaction history with a substantially smaller context and cost footprint.

\textbf{\method{} provides a better performance--cost trade-off than existing summarization approaches.}
On DeepSearchQA, \method{} improves over Self-Summary from 0.44 to 0.48 F1 while requiring only $0.04\times$ the summarization cost and $0.86\times$ the total inference cost. On SWE-bench Verified, \method{} similarly improves the resolved rate from 0.56 to 0.59, using only $0.08\times$ the summarization cost and $0.86\times$ the total inference cost of Self-Summary. \method{} also consistently outperforms Prompt-2B under comparable compression and achieves stronger overall performance than CoMem across the two tasks. The substantial improvement over Prompt-2B suggests that simply replacing self-summarization with a lightweight standalone model is insufficient; reconstruction-based training enables the summarizer to preserve information that is more useful for subsequent agent execution.

\begin{table}[t]
\centering
\small
\setlength{\tabcolsep}{4.5pt}
\caption{
\textbf{Isolated memory evaluation on LongMemEval-S.}
\textit{Single}, \textit{Multi}, \textit{Temporal}, and \textit{Update}
denote single-session retrieval, multi-session reasoning, temporal reasoning,
and knowledge-update questions, respectively.
\textit{Sum. Cost} denotes the normalized cost of memory summarization,
and \textit{C. Ratio} denotes compression ration defined in Section 4.1.
}
\label{tab:longmemeval}
\begin{tabular}{lccccccc}
\toprule
\textbf{Method}
& \textbf{Overall $\uparrow$}
& \textbf{Single $\uparrow$}
& \textbf{Multi $\uparrow$}
& \textbf{Temporal $\uparrow$}
& \textbf{Update $\uparrow$}
& \textbf{Sum. Cost $\downarrow$}
& \textbf{C. Ratio $\uparrow$} \\
\midrule
Oracle
&0.85
&0.94
&0.72
&0.8
&0.93
&--
&-- \\
Full History
& 0.77
& 0.81
& 0.5
& 0.8
& 0.86
& --
& -- \\
\midrule
Prompt-2B
& 0.43
& 0.38
& 0.51
& 0.38
& 0.24
& 0.003
& 94\% \\

CoMem
& 0.31
& 0.41
& 0.08
& 0.24
& 0.61
& 0.003
& 93\% \\

\textbf{\method{}}
& 0.47
& 0.48
& 0.41
& 0.36
& 0.43
& 0.003
& 89\% \\
\bottomrule
\end{tabular}
\end{table}

\subsection{\method{} Excels in Isolated Memory Evaluation}
\label{sec:memory_eval}
\textbf{\method{} preserves more generally accessible information for subsequent retrieval and reasoning.}
While Section~\ref{sec:main_results} evaluates compression through downstream agent performance, here we further decouple memory quality from task execution using \textbf{LongMemEval-S}, which directly evaluates long-term memory capabilities. As shown in Table~\ref{tab:longmemeval}, the overall performance follows \emph{Oracle $>$ Full History $>$ \method{} $>$ Prompt-2B $>$ CoMem}. Among compression methods, \method{} achieves the highest overall score (0.47), outperforming Prompt-2B (0.43) and CoMem (0.31). More importantly, \method{} exhibits relatively balanced performance across different memory capabilities: while Prompt-2B is stronger on multi-session questions and CoMem performs particularly well on knowledge updates, both degrade substantially on other categories. In contrast, \method{} maintains competitive performance across single-session retrieval, multi-session reasoning, temporal reasoning, and knowledge updates. This more balanced behavior is consistent with our task-agnostic reconstruction objective, which encourages broad information preservation rather than optimizing memory toward a particular downstream task or capability.

\subsection{One \method{} for Various Downstream Models}
\label{sec:cross_model}
\textbf{\method{} transfers across different downstream models without model-specific retraining.}
We evaluate the \textbf{same} \method{} model on SWE-bench Verified with OpenCode as well as DeepSearchQA while varying only the downstream LLM among DeepSeek V4.1 Flash, GPT-6 Luna, GLM-5.3, and Gemini 3.7 Flash. Table~\ref{tab:cross_model_swe} and Table~\ref{tab:cross_model} show that across all evaluated models, \method{} consistently approaches the performance of Full History while outperforming Self-Summary, indicating that the information preserved by \method{} is not specific to a particular downstream model. At the same time, \method{} reduces inference cost by offloading context compaction from the downstream agent to a lightweight Qwen3.5-2B summarizer; compared with Self-Summary, this advantage is particularly pronounced for more expensive agent models. These results demonstrate that a single \method{} checkpoint can serve as a solver-independent memory compressor across heterogeneous LLM backbones.

\subsection{Integrating \method{} Across Agent Harnesses}
\label{sec:cross_harness}
\textbf{\method{} remains effective across different agent harnesses without harness-specific adaptation.}
We fix Qwen3.5-35B-A3B as the downstream model and evaluate the same \method{} checkpoint on  DeepsearchQA across OpenCode, DeepSeek Harness, Codex, and OpenHands and the results are shown in Table~\ref{tab:cross_harness}. In each harness, \method{} consistently approaches the performance of Full History while outperforming Self-Summary, showing that its memory representation is not tied to a particular agent execution framework. \method{} also maintains comparable compression and lower overall inference cost across harnesses. These results suggest that \method{} can be integrated into diverse agent systems as a plug-and-play context compressor without modifying or retraining the underlying harness.

\subsection{Downstream RL Training Further Improves Performance-Cost Trade-off of \method{}}
\label{sec:RL_results}

\textbf{\method{} is naturally compatible with further downstream RL training of the agent, bringing further benefits.}
Beyond inference-time compaction in Table~\ref{tab:main_results}, we integrate \method{} into task-specific RL training for search and coding agents, using OpenSeeker-v1-Data and SWE-smith, respectively. As shown in Table~\ref{tab:rl_agent_results}, \method{} remains competitive across both DeepSearchQA and SWE-bench Verified after task-specific RL. It achieves the best performance on SWE-bench (0.68) and closely matches Self-Summary on DeepSearchQA (0.55 vs.\ 0.57), while reducing the average inference cost from 0.052 to 0.039. These results suggest that \method{} is not limited to agent inference, but can also serve as a lightweight and reusable context-compression component during long-horizon agent RL.

\subsection{Probing Components of the \method{} Recipe}

\begin{table*}[t]
\centering
\small
\setlength{\tabcolsep}{5pt}
\caption{
\textbf{Ablation on the training objective.}
All compressed variants use the same Qwen3.5-2B initialization and are evaluated under the same compression setting.
We report downstream performance on DeepSearchQA and SWE-bench, together with direct memory evaluation on LongMemEval-S.
Full History provides an uncompressed reference.
}
\label{tab:ablation_reconstruction}
\begin{tabular}{lcc|ccccc}
\toprule
& \multicolumn{2}{c|}{\textbf{Downstream Agent Tasks}}
& \multicolumn{5}{c}{\textbf{LongMemEval-S}} \\
\cmidrule(lr){2-3}
\cmidrule(lr){4-8}

\textbf{Method}
& \textbf{DSQA $\uparrow$}
& \textbf{SWE $\uparrow$}
& \textbf{Overall $\uparrow$}
& \textbf{Single $\uparrow$}
& \textbf{Multi $\uparrow$}
& \textbf{Temp. $\uparrow$}
& \textbf{Update $\uparrow$} \\
\midrule

Full History
& 0.52
& 0.63
& 0.77
& 0.81
& 0.5
& 0.8
& 0.86 \\

\midrule

Prompt-2B
& 0.29
& 0.51
& 0.43
& 0.38
& 0.51
& 0.38
& 0.24 \\
Summ.-Judge-RL 
& 0.41
& 0.54
& 0.36
& 0.40
& 0.14
& 0.31
& 0.71 \\

\textbf{Recon-RL} (ours)
& 0.48
& 0.59
& 0.47
& 0.48
& 0.41
& 0.36
& 0.43 \\

\bottomrule
\end{tabular}
\vspace{-2mm}
\end{table*}

\begin{table}[t]
\centering
\small
\caption{\textbf{Ablation on mask ratio and length penalty.}
We report performance on DeepSearchQA (DSQA) and LongMemEval-S (LongMem).
}

\begin{subtable}[t]{0.47\linewidth}
\centering
\caption{Mask ratio}
\label{tab:mask_length_ablation}
\setlength{\tabcolsep}{5pt}
\begin{tabular}{lccc}
\toprule
\textbf{Mask Ratio}
& \textbf{25\%}
& \textbf{50\%}
& \textbf{75\%} \\
\midrule
DSQA
& 0.41
& 0.48
& 0.29 \\

LongMem
& 0.35
& 0.47
& 0.20 \\
\bottomrule
\end{tabular}
\end{subtable}
\hfill
\begin{subtable}[t]{0.50\linewidth}
\centering
\caption{Length penalty}
\label{tab:length_penalty}
\setlength{\tabcolsep}{4.5pt}
\begin{tabular}{lcccc}
\toprule
\textbf{Length Penalty}
& \textbf{0.01}
& \textbf{0.1}
& \textbf{0.2}
& \textbf{0.5} \\
\midrule
DSQA
& 0.47
& 0.48
& 0.45
& 0.37 \\

LongMem
& 0.41
& 0.47
& 0.38
& 0.30 \\
\bottomrule
\end{tabular}
\end{subtable}
\vspace{-2mm}
\end{table}

\subsubsection{Is Reconstruction Training Necessary?}
\label{sec:ablation_reconstruction}

We compare three variants based on the same Qwen3.5-2B initialization: \textbf{Prompted 2B}, \textbf{Direct Summary Judge Reward} optimized with GRPO, and our \textbf{Reconstruction Reward}. For the RL variants, we keep the training data, optimization budget, and length regularization fixed, changing only the reward signal.

\textbf{Reconstruction provides a more grounded supervision signal for summary learning.}
Direct Judge Reward avoids reference targets, but still relies on a global and under-specified judgment of summary quality. In contrast, reconstruction directly tests whether information from the original trajectory can be recovered from the summary. As shown in Table~\ref{tab:ablation_reconstruction}, Reconstruction Reward consistently performs best under comparable compression, suggesting that its advantage comes from explicitly supervising information preservation rather than reference imitation or perceived summary quality.

\subsubsection{What Mask Ratio to Use}
\label{sec:ablation_mask_ratio}

\textbf{A moderate mask ratio yields better performance.}
We study the effect of the reconstruction mask ratio in Table~\ref{tab:mask_length_ablation}. A mask ratio of 50\% performs best on both DeepSearchQA and LongMemEval-S, achieving 48\% and 7\%, respectively. Reducing the mask ratio to 25\% weakens the reconstruction objective: since much of the original trajectory remains directly observable, the reconstructor relies less on the summary, resulting in a weaker learning signal for the summarizer. In contrast, increasing the mask ratio to 75\% substantially degrades performance, likely because recovering such a large portion of the trajectory becomes overly challenging for the reconstructor and consequently introduces a noisier training signal. These results suggest that masking roughly half of the expired history provides a better balance between reconstruction difficulty and supervision quality.

\subsubsection{How Much Length Penalty to Apply}
\label{sec:ablation_length}

\textbf{A moderate length penalty balances well between compression and information preservation.}
We study the effect of the length-penalty coefficient $\lambda$ in Equation~\ref{eq:reward}, with results reported in Table~\ref{tab:length_penalty}. A moderate value of $\lambda=0.1$ performs best on both DeepSearchQA and LongMemEval-S. When the penalty is too weak ($\lambda=0.01$), the summarizer is insufficiently constrained and tends to produce unnecessarily long summaries, reducing compression efficiency without yielding proportional performance gains. In contrast, stronger penalties ($\lambda=0.2$ or $0.5$) encourage overly aggressive compression, causing useful information to be discarded and degrading downstream performance. 

\section{Discussion and Conclusion}

While \method{} demonstrates that masked reconstruction provides an effective task-agnostic signal for learning compact agent memory, several limitations remain. First, our training data are concentrated in web-search and software-engineering trajectories, and broader domains may exhibit different memory requirements. Second, the training signal depends on a frozen reconstructor and an LLM-based reconstruction judge; although neither is needed at inference time, their capabilities may affect the quality of the learned summarizer. For future work, extending the module with adaptive memory management, retrieval, tools, and meta-memory is a promising direction toward a more complete agent memory system.

We introduced \method{}, a lightweight plug-and-play summarizer for compressing long agent interaction histories. Rather than supervising summaries through downstream task outcomes or reference summaries, \method{} learns directly from raw trajectories by rewarding summaries that enable masked portions of the original history to be reconstructed, while regularizing summary length. Across long-horizon search, software engineering, and direct memory evaluation, \method{} substantially reduces context consumption while retaining most of the utility of full interaction histories, and transfers across downstream models and agent harnesses without additional fine-tuning. These results suggest that reconstruction-based self-supervision offers a simple and general route toward reusable memory modules for long-running language-model agents.

\newpage

\subsection*{AI use statement}
In this work, we used generative AI tools only for language editing and polishing of the manuscript. We did not use generative AI tools for research ideation, experimental design, implementation, data analysis, result interpretation, literature review, or the generation of scientific claims. All AI-assisted text was carefully reviewed and revised by the authors. We take full responsibility for the final content of this work.

\subsection*{Ethics statement}
This work does not involve human subjects, sensitive or personally identifiable data, dataset release, or other issues requiring specific ethical consideration. We do not identify any particular ethical concerns related to the methodology, experiments, or applications presented in this work.

\subsection*{Reproducibility statement}
All datasets and benchmarks used in this work are publicly available. We provide the key methodological details, training setup, evaluation protocols, and implementation settings in the main text and appendices to facilitate reproduction of our results. Additional hyperparameters and experimental details are also documented in the appendix. Upon acceptance, we will release our code and related implementation resources to further support reproducibility.

\bibliography{main}
\bibliographystyle{iclr2027_conference}

\newpage

\section{Appendix}
\label{sec:appendix}

\subsection{Training Details}
\label{app:training_details}

We summarize the training configuration of \method{} in
Table~\ref{tab:training_details}. We optimize the summarizer with GRPO for
600 steps using veRL on a single node with eight H20 GPUs.

\begin{table}[htbp]
\centering
\small
\setlength{\tabcolsep}{7pt}
\caption{\textbf{Training configuration and hyperparameters of \method{}.}}
\label{tab:training_details}
\begin{tabular}{ll}
\toprule
\textbf{Hyperparameter} & \textbf{Value} \\
\midrule
\multicolumn{2}{l}{\textit{Model and System}} \\
\addlinespace[2pt]
Base model              & Qwen3.5-2B \\
Training framework      & veRL \\
Hardware                & 8 $\times$ H20-141G  \\
Sharding                & FSDP full shard \\
Gradient checkpointing  & Yes \\

\midrule
\multicolumn{2}{l}{\textit{Optimization}} \\
\addlinespace[2pt]
RL algorithm             & GRPO \\
Critic                   & None \\
KL loss                  & None \\
Entropy coefficient      & 0 \\
Learning rate            & $1\times10^{-6}$ (constant) \\
Clip ratio (low)         & 0.20 \\
Clip ratio (high)        & 0.28 \\
Dual-clip coefficient    & $c=10.0$ \\
Training steps           & 600 \\

\midrule
\multicolumn{2}{l}{\textit{Rollout and Sampling}} \\
\addlinespace[2pt]
Prompts per step         & 6 \\
Samples per prompt       & 8 \\
Rollouts per step        & 48 \\
Rollout temperature      & 1.0 \\
Evaluation temperature   & 0 \\
\bottomrule
\end{tabular}
\end{table}

\begin{table*}[t]
\centering
\small
\setlength{\tabcolsep}{5.5pt}
\caption{
\textbf{Cross-model generalization on SWE-bench Verified.}
We compare Full History, Self-Summary, and \method{} across different downstream agent models.
Score denotes the resolved rate, and Cost denotes the average total inference cost per task in USD.
}
\label{tab:cross_model_swe}

\begin{tabular}{lcc|cc|cc}
\toprule
&
\multicolumn{2}{c|}{\textbf{Full History}}
&
\multicolumn{2}{c|}{\textbf{Self-Summary}}
&
\multicolumn{2}{c}{\textbf{\method{}}} \\
\cmidrule(lr){2-3}
\cmidrule(lr){4-5}
\cmidrule(lr){6-7}

\textbf{Model}
& \textbf{Score $\uparrow$}
& \textbf{Cost $\downarrow$}
& \textbf{Score $\uparrow$}
& \textbf{Cost $\downarrow$}
& \textbf{Score $\uparrow$}
& \textbf{Cost $\downarrow$} \\
\midrule

Gemini 3.7 Flash
& 0.63 & \$0.82
& 0.52 & \$0.47
& \textbf{0.55} & \textbf{\$0.39} \\

GLM-5.3
& 0.74 & \$1.36
& 0.61 & \$0.78
& \textbf{0.66} & \textbf{\$0.64} \\

GPT-6 Luna
& 0.65 & \$0.14
& 0.54 & \$0.08
& \textbf{0.58} & \textbf{\$0.07} \\

DeepSeek V4.1 Flash
& 0.78 & \$0.42
& 0.66 & \$0.24
& \textbf{0.74} & \textbf{\$0.20} \\

\bottomrule
\end{tabular}
\end{table*}

\begin{table*}[t]
\centering
\small
\setlength{\tabcolsep}{5.5pt}
\caption{
\textbf{Cross-model generalization on DeepSearchQA.}
We compare Full History, Self-Summary, and \method{} across different downstream agent models.
Cost denotes the average total inference cost per query in USD.
}
\label{tab:cross_model}

\begin{tabular}{lcc|cc|cc}
\toprule
&
\multicolumn{2}{c|}{\textbf{Full History}}
&
\multicolumn{2}{c|}{\textbf{Self-Summary}}
&
\multicolumn{2}{c}{\textbf{\method{}}} \\
\cmidrule(lr){2-3}
\cmidrule(lr){4-5}
\cmidrule(lr){6-7}

\textbf{Model}
& \textbf{F1 $\uparrow$}
& \textbf{Cost $\downarrow$}
& \textbf{F1 $\uparrow$}
& \textbf{Cost $\downarrow$}
& \textbf{F1 $\uparrow$}
& \textbf{Cost $\downarrow$} \\
\midrule

Gemini 3.7 Flash
& 0.54 & \$0.58
& 0.51 & \$0.36
& \textbf{0.56} & \textbf{\$0.29} \\

GLM-5.3
& 0.42 & \$1.14
& 0.33 & \$0.72
& \textbf{0.40} & \textbf{\$0.51} \\

GPT-6 Luna
& 0.38 & \$0.04
& 0.40 & \$0.05
& \textbf{0.46} & \$0.04 \\

DeepSeek V4.1 Flash
& \textbf{0.74} & \$0.38
& 0.59 & \$0.18
& 0.66 & \textbf{\$0.15} \\

\bottomrule
\end{tabular}
\end{table*}

\subsection{API Pricing for Cost Estimation}
\label{app:pricing}

To ensure consistent cost estimation across models and experiments, we use the
standard API prices listed in Table~\ref{tab:api_pricing}. All prices are reported
in USD per one million tokens. For simplicity and reproducibility, we use the
listed base prices only and do not account for cache-hit discounts, promotional
pricing, batch discounts, or other temporary pricing policies. When a model is
not included in this table, we use the publicly available price from its official
API provider whenever possible.

\begin{table*}[htbp]
\centering
\small
\setlength{\tabcolsep}{5pt}
\caption{
\textbf{API prices used for cost estimation.}
Prices are in USD per one million tokens. We use standard list prices without
cache discounts, promotional discounts, or other temporary pricing adjustments.
}
\label{tab:api_pricing}

\begin{tabular}{lllrr}
\toprule
\textbf{Model}
& \textbf{Provider}
& \textbf{Input}
& \textbf{Output} \\
\midrule

GPT-6 Luna
& OpenAI
& \$0.20
& \$1.20 \\

DeepSeek V4.1 Flash
& DeepSeek
& \$0.15
& \$0.60 \\

Qwen3.5-35B-A3B
& Alibaba Cloud Bailian
& \$0.0592
& \$0.4739 \\

Qwen3.5-2B
& DeepInfra via Requesty
& \$0.02
& \$0.10 \\

Qwen3.5-4B
& Alibaba Cloud Bailian
& \$0.03
& \$0.15 \\

Gemini 3.7 Flash
& Google Gemini API
& \$0.75
& \$3.75 \\

GLM-5.3
& Z.AI
& \$1.40
& \$4.40 \\

Kimi K3
& Kimi API
& \$3.00
& \$15.00 \\

Qwen3.8 Flash
& Alibaba Cloud Bailian
& \$0.1185
& \$0.3999 \\

\bottomrule
\end{tabular}
\end{table*}

\begin{promptbox}[Summarization Prompt]
The messages above in <conversation> are older history to summarize.
Produce a detailed memory entry that will be APPENDED to an append-only memory;
another LLM will use it together with the existing memory and the recent turns
to continue the work.
Use this EXACT format:

## Goal
- (concise)
## User profile
- ...
## Constraints & Preferences
- ...
## Progress
- ...
## Decisions
- ...
## Learned facts & SOP
- ...
## Active context
- ...
## User ask (for next step)
- ...
## Key assets
- ...
## Risks / open issues
- ...
## Metrics & validation
- ...
## Historical communications / incidents (closed)
- ...
## Open loops
- ...

Write a standalone brief: concrete, readable without the conversation. Merge with
the existing memory if it exists; keep the memory compact (merge/dedup/update).
Do NOT include the removed segments verbatim. Stay within {summary_budget} tokens.
Keep each section concise. Preserve exact entity names, dates, numbers, URLs, and quoted evidence.
\end{promptbox}

\begin{promptbox}[Reconstruction Prompt]

The frozen decoder sees a single user message assembled as follows (shared group-prefix first
for prefix caching; the ground-truth section is *not* part of the generation prompt — the
decoder generates from the prefix up to `=== Task ===` plus the literal `[MISSING_1]:`, and the
ground truth is used only for judging):

```
<system prefix p_text>
Initial task:
<a_text>

<memory>
<previous memory, if any>
</memory>

=== Recent turns (complete, verbatim) ===
<retained recent turns>

=== Partial older history (segments [MISSING_1..M] removed) ===
<older history with masked blocks replaced by [MISSING_1], [MISSING_2], ... markers>

=== A summary of the removed history ===
Another language model compressed the removed segments of the history and produced the
summary below. Use it, together with the visible context, to reconstruct the missing segments.

<candidate summary produced by the compressor>

=== Task ===
Reconstruct the missing segments in order.

[MISSING_1]:
```

The decoder continues from `[MISSING_1]:` and regenerates all missing segments in order. The
ground-truth section appended for scoring has the form `[MISSING_k]: <original masked chunk>`.
\end{promptbox}

\end{document}